\documentclass{article}
\usepackage{iclr2026_conference,times}

\usepackage{amsmath,amsfonts,bm}

\def\eqref#1{equation~\ref{#1}}

\def\1{\bm{1}}

\DeclareMathAlphabet{\mathsfit}{\encodingdefault}{\sfdefault}{m}{sl}
\SetMathAlphabet{\mathsfit}{bold}{\encodingdefault}{\sfdefault}{bx}{n}

\usepackage{hyperref}
\usepackage{url}
\usepackage{booktabs}
\usepackage{amsmath,amssymb,amsthm}
\usepackage{tikz}
\usetikzlibrary{positioning,arrows.meta}

\iclrfinalcopy

\title{DeltaLogic: Minimal Premise Edits Reveal Belief-Revision Failures in Logical Reasoning Models}
\author{\textbf{Amit Dhanda}\thanks{This work does not relate to the author's position at Amazon.}\\
Amazon\\
\texttt{amdhanda@amazon}}

\newtheorem{definition}{Definition}

\begin{document}

\maketitle
\lhead{Accepted at ICLR 2026 Workshop on Logical Reasoning of Large Language Models}

\begin{abstract}
Reasoning benchmarks typically evaluate whether a model derives the correct answer from a fixed premise set, but they under-measure a closely related capability that matters in dynamic environments: \emph{belief revision under minimal evidence change}. We introduce \textbf{DeltaLogic}, a benchmark transformation protocol that converts natural-language reasoning examples into short revision episodes. Each episode first asks for an initial conclusion under premises $P$, then applies a minimal edit $\delta(P)$, and finally asks whether the previous conclusion should remain stable or be revised. We instantiate DeltaLogic from FOLIO and ProofWriter and evaluate small causal language models with constrained label scoring. On a completed 30-episode Qwen evaluation subset, stronger initial reasoning still does not imply stronger revision behavior: Qwen3-1.7B reaches $0.667$ initial accuracy but only $0.467$ revision accuracy, with inertia rising to $0.600$ on episodes where the gold label should change, while Qwen3-0.6B collapses into near-universal abstention. There, Qwen3-4B preserves the same inertial failure pattern ($0.650$ initial, $0.450$ revised, $0.600$ inertia), whereas Phi-4-mini-instruct is substantially stronger ($0.950$ initial, $0.850$ revised) but still exhibits non-trivial abstention and control instability. These results suggest that logical competence under fixed premises does not imply disciplined belief revision after local evidence edits. DeltaLogic therefore targets a distinct and practically important reasoning capability that complements existing logical inference and belief-updating benchmarks.
\end{abstract}

\section{Introduction}
Reasoning benchmarks usually evaluate inference from fixed premises. DeltaLogic targets a different capability: \emph{local belief revision}. When one premise is inserted, deleted, or replaced, the model should update exactly the commitments that the edit warrants and leave the rest stable. This matters for systems that reason over changing documents, dynamic rules, or incrementally updated evidence.

DeltaLogic turns a standard reasoning item into a short revision episode with a known semantic effect. This yields a failure taxonomy that static accuracy cannot expose: \emph{inertia} (keeping an outdated answer), \emph{over-flip} (revising under an irrelevant edit), and \emph{degenerate abstention}. Our contributions are a simple benchmark-construction protocol, metrics that separate initial reasoning from revision discipline, and generative-model evidence that belief revision is a distinct challenge even for current small and near-4B models.

This paper is a \emph{measurement} contribution, not a new reasoning algorithm. The claim is not that DeltaLogic explains the internal mechanism of belief revision in language models. The claim is narrower and testable: current small LMs that appear competent on static reasoning still fail under minimal premise edits, and those failures can be decomposed into stable, measurable revision modes.

\paragraph{Contributions.}
(1) We introduce a benchmark-transformation protocol for local belief revision over public logical reasoning datasets. (2) We define a small set of revision-specific metrics that separate stale commitment, unnecessary revision, and abstention. (3) We show on completed runs that stronger static reasoning does not guarantee better local revision, and that models of similar scale can fail through different revision modes.

\section{Methodology}
Let an original reasoning instance consist of premises $P$, query $q$, and gold label $y_0$. A DeltaLogic episode applies a minimal edit operator $\delta$ to obtain revised premises $P'=\delta(P)$ with revised gold label $y_1$. The edit is designed so that its intended semantic effect is deterministic.

\begin{definition}[DeltaLogic episode]
A DeltaLogic episode is a tuple
\[
e=(P,q,y_0,P',y_1,t),
\]
where $P$ is the original premise set, $q$ is the hypothesis or query, $y_0$ is the original gold label, $P'$ is the minimally edited premise set, $y_1$ is the revised gold label, and $t$ is the edit type.
\end{definition}

We use four edit types: support insertion, defeating-fact insertion, support removal, and irrelevant-fact addition. Together they test positive updating, belief retraction, and stability under no-change controls.

\begin{figure}[t]
\centering
\begin{tikzpicture}[
  node distance=6mm and 8mm,
  >=Latex,
  box/.style={draw, rounded corners=1.5mm, align=center, minimum width=3.2cm, minimum height=0.9cm},
  arr/.style={-Latex, thick}
]
\node[box] (source) {Source example\\premises $P$, query $q$, label $y_0$};
\node[box, right=of source] (edit) {Minimal edit\\$\delta(P)$};
\node[box, right=of edit] (revised) {Revised state\\$P'$, label $y_1$};
\node[box, below=of edit] (metrics) {Metrics\\init/revision acc.\\inertia, over-flip, abstain};
\draw[arr] (source) -- node[above, font=\scriptsize] {construct} (edit);
\draw[arr] (edit) -- node[above, font=\scriptsize] {apply} (revised);
\draw[arr] (source.south) |- (metrics.west);
\draw[arr] (revised.south) |- (metrics.east);
\end{tikzpicture}
\caption{DeltaLogic construction pipeline. A standard reasoning item is turned into a minimally edited revision episode with a known semantic effect, which makes revision errors measurable rather than anecdotal.}
\label{fig:pipeline}
\end{figure}
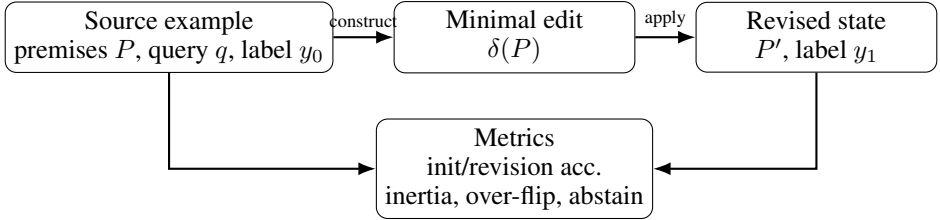

DeltaLogic is instantiated from FOLIO and ProofWriter. From FOLIO, we build support-insertion, defeating-fact, and irrelevant-addition episodes. From ProofWriter, we build support-removal and irrelevant-addition episodes using shallow examples with identified support facts. The full construction contains 100 episodes. For this paper we report completed generative-model evaluation subsets that preserve all edit types while remaining feasible on CPU-only hardware: a 30-episode main Qwen subset and a 20-episode near-4B extension.

\section{Metrics}
We report initial accuracy,
\[
\mathrm{Acc}_{\mathrm{init}} = \Pr(\hat y_0 = y_0),
\]
and revision accuracy,
\[
\mathrm{Acc}_{\mathrm{rev}} = \Pr(\hat y_1 = y_1).
\]
To expose failure modes, let $C$ denote episodes whose gold label should change and $U$ denote episodes whose gold label should remain stable. The inertia rate is
\[
\mathrm{Inertia}
=
\Pr(\hat y_1 = \hat y_0 \mid e \in C,\ \hat y_0 = y_0),
\]
The over-flip rate is
\[
\mathrm{OverFlip}
=
\Pr(\hat y_1 \neq \hat y_0 \mid e \in U,\ \hat y_0 = y_0),
\]
We also report abstention rate,
\[
\mathrm{Abstain}=\Pr(\hat y_1=\texttt{Uncertain}),
\]
because some models appear stable only by retreating into uncertainty.

We also refer to the empirical \emph{revision gap}, $\mathrm{Acc}_{\mathrm{init}}-\mathrm{Acc}_{\mathrm{rev}}$, as a compact measure of how much performance degrades after the evidence edit.

\section{Experimental Setup}
We evaluate Qwen3-0.6B, Qwen3-1.7B, Qwen3-4B, and Phi-4-mini-instruct. Prompts end in \texttt{Label:}, and we score the continuations \texttt{True}, \texttt{False}, and \texttt{Uncertain} by average token log-likelihood under the frozen causal LM. This constrained scoring avoids chain-of-thought parsing artifacts and uses no task-specific fine-tuning or verifier model. The completed experiments are split into two evaluation subsets: a 30-episode Qwen main study and a 20-episode near-4B extension. We keep them separate in reporting because the subsets are not identical.

\section{Results}
Table~\ref{tab:main-results} reports all completed runs. The central finding is that stronger initial reasoning still does not imply stronger revision behavior. On the 30-episode Qwen evaluation subset, Qwen3-1.7B reaches $0.667$ initial accuracy but only $0.467$ revision accuracy, a revision gap of $0.200$, while Qwen3-0.6B reaches $0.400$ revision accuracy only through universal abstention. On the completed 20-episode near-4B evaluation subset, Qwen3-4B preserves the same inertial failure pattern, whereas Phi-4-mini-instruct is substantially stronger but not stable.

\begin{table}[t]
\caption{Main DeltaLogic results on the completed generative-model evaluation subsets.}
\label{tab:main-results}
\centering
\begin{tabular}{lcccccc}
\toprule
Model & Slice $n$ & Init. $\uparrow$ & Rev. $\uparrow$ & Inertia $\downarrow$ & Over-flip $\downarrow$ & Abstain $\downarrow$ \\
\midrule
Qwen3-0.6B & 30 & 0.300 & 0.400 & 0.400 & 0.000 & 1.000 \\
Qwen3-1.7B & 30 & 0.667 & 0.467 & 0.600 & 0.000 & 0.000 \\
Qwen3-4B & 20 & 0.650 & 0.450 & 0.600 & 0.000 & 0.000 \\
Phi-4-mini & 20 & 0.950 & 0.850 & 0.200 & 0.100 & 0.350 \\
\bottomrule
\end{tabular}
\end{table}

The edit-type breakdown reveals the real structure of the problem. Qwen3-1.7B succeeds on support-insertion episodes with perfect revision accuracy and remains perfectly stable on the ProofWriter irrelevant-addition control. Yet it fails completely on support-removal and defeating-fact edits. This asymmetry is scientifically important. Adding a new premise that explicitly supports the query is easier than revising a previously justified belief once its support disappears or once a defeating fact arrives. The harder case requires reasoning about the loss or reversal of support rather than simply reacting to explicit positive evidence.

Qwen3-0.6B exhibits a different pathology. Its revised abstention rate is $1.000$, so its apparent stability is not genuine revision discipline. It gets support-removal episodes correct only because the gold revised label in that regime is also \texttt{Uncertain}. The smaller model therefore avoids stale commitments by retreating into blanket uncertainty, whereas the larger Qwen models preserve sharper beliefs but fail to revise them when they should change. Qwen3-4B does not materially improve over Qwen3-1.7B on inertia.

Phi-4-mini-instruct behaves differently. It is the strongest completed model in the paper, but it still shows a non-zero over-flip rate and a substantial abstention rate. The main scientific takeaway is therefore not simply that a stronger model solves the problem; rather, DeltaLogic distinguishes between multiple revision regimes. The Qwen family is primarily inertial, while Phi-4-mini is more revision-capable but still hedges and occasionally revises unnecessarily. The edit breakdown makes this concrete: Qwen3-4B remains at $0.000$ revision accuracy on ProofWriter support-removal episodes, while Phi-4-mini reaches $0.667$ on the same edit type.

The important observation is not just that one model scores higher. It is that scale alone does not induce the right \emph{local update rule}. Qwen3-4B is larger than Qwen3-1.7B but preserves essentially the same inertia profile. That is evidence against a naive ``more scale fixes revision'' story and in favor of the paper's narrower claim: minimal belief revision is a distinct capability worth measuring directly.

\begin{figure}[t]
\centering
\begin{tikzpicture}[x=5.2cm,y=0.72cm]
\tikzset{lab/.style={font=\scriptsize}, small/.style={font=\tiny}}
\node[lab, anchor=east] at (-0.05,0) {Qwen3-0.6B};
\node[lab, anchor=east] at (-0.05,1) {Qwen3-1.7B};
\node[lab, anchor=east] at (-0.05,2) {Qwen3-4B};
\node[lab, anchor=east] at (-0.05,3) {Phi-4-mini};
\foreach \x/\lbl in {0/0.0,0.5/0.5,1.0/1.0} {
  \draw[gray!40] (\x,-0.45) -- (\x,3.45);
  \node[small, anchor=north] at (\x,-0.48) {\lbl};
}
\fill[blue!60] (0,-0.22) rectangle (0.40,-0.04);
\fill[orange!80] (0,-0.02) rectangle (0.00,0.16);
\fill[red!70] (0,0.18) rectangle (1.00,0.36);
\fill[blue!60] (0,0.78) rectangle (0.60,0.96);
\fill[orange!80] (0,0.98) rectangle (0.00,1.16);
\fill[red!70] (0,1.18) rectangle (0.00,1.36);
\fill[blue!60] (0,1.78) rectangle (0.60,1.96);
\fill[orange!80] (0,1.98) rectangle (0.00,2.16);
\fill[red!70] (0,2.18) rectangle (0.00,2.36);
\fill[blue!60] (0,2.78) rectangle (0.20,2.96);
\fill[orange!80] (0,2.98) rectangle (0.10,3.16);
\fill[red!70] (0,3.18) rectangle (0.35,3.36);
\node[small, anchor=west, text=blue!60!black] at (0.02,3.62) {Inertia};
\node[small, anchor=west, text=orange!90!black] at (0.32,3.62) {Over-flip};
\node[small, anchor=west, text=red!70!black] at (0.72,3.62) {Abstain};
\end{tikzpicture}
\caption{Failure-mode comparison across completed runs. The Qwen family remains inertia-dominated, while Phi-4-mini trades lower inertia for non-zero over-flip and abstention.}
\label{fig:failure-modes}
\end{figure}
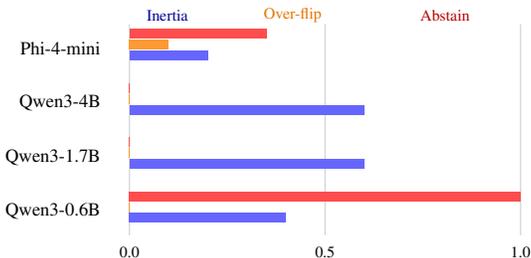

\section{Related Work}
DeltaLogic is closest to three neighboring lines. First, Belief-R studies adaptation under new evidence and motivates the general update-versus-stability tension \citep{beliefr}. DeltaLogic differs by using smaller local edits with deterministic semantic effects. Second, logical-reasoning benchmarks such as FOLIO, ProofWriter, LogicBench, and LogicGame measure inference from fixed premises \citep{folio,proofwriter,logicbench,logicgame}. DeltaLogic is layered on top of them as a transformation protocol for revision rather than static inference. Third, revision-oriented question answering and correction benchmarks such as ReviseQA study answer updating under changing evidence, but do not focus on logically controlled minimal premise edits. DeltaLogic asks a narrower question: after one evidence edit, does the model update exactly the commitments that should change and keep the rest stable?

\section{Limitations}
DeltaLogic is a controlled benchmark built from transformed FOLIO and ProofWriter examples rather than open-world interactive agents, so the results trade ecological breadth for precise edit semantics. The current submission also reports completed but unevenly sized model evaluations because causal-LM scoring on the available CPU-only setup is expensive. Accordingly, the results support the existence and structure of the failure mode rather than a definitive leaderboard. Finally, the benchmark targets single-step revision and label correctness, not longer revision chains or free-form justification quality.

\section{Conclusion}
Belief revision under minimal premise edits is a distinct reasoning capability that current evaluations only partially capture. DeltaLogic makes that capability measurable by converting standard reasoning instances into short revision episodes with known semantic effects. Across the completed runs in this paper, small causal LMs fail in at least three distinct ways: clinging to stale commitments, collapsing into uncertainty, or revising mostly correctly but with residual abstention and occasional over-flip. The practical implication is straightforward: getting the original answer right is not enough. A reliable reasoning model must also know how to update that answer precisely when the evidence changes, and DeltaLogic provides a direct probe of that ability.

\appendix
\section{Appendix: Algorithmic Details}

\paragraph{Benchmark construction.}
Each DeltaLogic episode is constructed from a base example $(P, q, y)$ with premise set $P$, query $q$, and gold label $y \in \{\texttt{True}, \texttt{False}, \texttt{Uncertain}\}$. We first select examples whose semantic status can be modified by one controlled edit. We then generate either a \emph{changed} episode, in which the label is intended to change, or a \emph{control} episode, in which the label should remain stable.

\begin{enumerate}
\item Sample a base example from FOLIO or ProofWriter.
\item Derive an edit operator $e$ from a fixed family: support insertion, defeating-fact insertion, support removal, or irrelevant-fact addition.
\item Apply $e$ to obtain revised premises $P' = e(P)$.
\item Compute the revised label $y'$ using dataset metadata and edit semantics.
\item Emit a four-turn episode:
  (i) answer the original query from $P$;
  (ii) optionally explain briefly which premise matters most;
  (iii) observe the revised premise set $P'$;
  (iv) answer the same query again and state whether the answer should change.
\end{enumerate}

\paragraph{Label-scoring inference.}
The current experiments use closed-label scoring rather than unconstrained generation. For each model, prompt state, and candidate label $a \in \{\texttt{True}, \texttt{False}, \texttt{Uncertain}\}$, we score the continuation by conditional log-likelihood,
\[
s(a \mid x) = \sum_{t=1}^{|a|} \log p_\theta(a_t \mid x, a_{<t}),
\]
where $x$ is the serialized dialogue context. The predicted label is
\[
\hat{a} = \arg\max_{a \in \{\texttt{True}, \texttt{False}, \texttt{Uncertain}\}} s(a \mid x).
\]
Initial accuracy is measured on the pre-edit state; revision accuracy is measured on the post-edit state. Inertia is the rate at which a model preserves the old label when the gold label changes. Over-flip is the rate at which a model changes its answer on control episodes where the gold label should remain stable. Abstention is the prediction rate for \texttt{Uncertain}.

\paragraph{Practical implementation.}
On causal language models, each label score requires a full forward pass over the dialogue prefix plus label tokens. This is why the generative-model experiments are more expensive than discriminative NLI baselines. All reported runs use deterministic preprocessing, fixed random seeds for benchmark construction, and locally cached model and dataset artifacts to avoid network variance.

\end{document}